%% file: main.tex
\documentclass{article}
\usepackage{spconf,amsmath,graphicx}
\usepackage{multirow, booktabs}
\usepackage{xcolor}
\usepackage{hyperref}
\usepackage{cleveref}

\title{Correction of automatic speech recognition \\ with transformer sequence-to-sequence model}

\name{Oleksii Hrinchuk$^{12*}$\thanks{*Equal contribution, work done during an internship at NVIDIA}\qquad
Mariya Popova$^{2*}$ \qquad Boris Ginsburg$^2$}
\address{$^1$Moscow Institute of Physics and Technology, Moscow, Russia \\ $^2$NVIDIA, Santa Clara, CA, USA \\
\texttt{\normalsize \{aleksey.grinchuk@phystech.edu, mariewelt@gmail.com, bginsburg@nvidia.com\}}}

\begin{document}
%
\maketitle
\begin{abstract}
In this work, we introduce a simple yet efficient post-processing model for automatic speech recognition (ASR). Our model has Transformer-based encoder-decoder architecture which ``translates'' ASR model output into grammatically and semantically correct text. We investigate different strategies for regularizing and optimizing the model and show that extensive data augmentation and the initialization with pre-trained weights are required to achieve good performance. On the LibriSpeech benchmark, our method demonstrates significant improvement in word error rate over the baseline acoustic model with greedy decoding, especially on much noisier dev-other and test-other portions of the evaluation dataset. Our model also outperforms baseline with 6-gram language model re-scoring and approaches the performance of re-scoring with Transformer-XL neural language model.
\end{abstract}

\begin{keywords}
speech recognition, spelling correction, pre-trained language models
\end{keywords}

\input{text/1_intro.tex}
\input{text/2_model.tex}
\input{text/3_experiments.tex}
\input{text/4_conclusion.tex}

\bibliographystyle{IEEEbib}
\bibliography{refs}

\end{document}

%% file: text/1_intro.tex
\section{Introduction}
\label{sec:intro}



In recent years, automatic speech recognition (ASR) research has been dominated by end-to-end (E2E) models~\cite{chan2016listen,bahdanau2016end,zhang2017very} which outperformed conventional hybrid systems relying on Hidden Markov Models~\cite{bengio1991global,hinton2012deep}. In contrast to prior work, which required training several independent components (acoustic, pronunciation, and language models) and had many degrees of complexity, E2E models are faster and easier to implement, train, and deploy.

To enhance the speech recognition accuracy, ASR models are often augmented with independently trained language models that re-score the list of n-best hypotheses. The use of the external language model induces a natural trade-off between model speed and accuracy. While simple N-gram language models (e.g., KenLM~\cite{heafield2011kenlm}) are extremely fast, they can not achieve the same level of performance as heavier and more powerful neural language models, such as Transformers~\cite{li2019,dai2019transformer,baevski2018adaptive}.

Language model re-scoring effectively expands the search space of speech recognition candidates; however, it can barely help when the ground truth word was assigned a low score by erroneous ASR model. Traditional left-to-right language models are also prone to error accumulation: if some word at the beginning of the decoded speech is misrecognized, it will affect the scores of all succeeding words by providing them with incorrect context. To address these problems, we propose to train a conditional language model that corrects the errors made by the system operating similar to neural machine translation (NMT)~\cite{sutskever2014sequence,cho2014learning} by ``translating'' corrupted ASR output into the correct language.

There is a plethora of prior work on correcting ASR systems output, and we refer the reader to~\cite{errattahi2018automatic} for a detailed overview. Most closely to our work, \cite{guo2019spelling} propose to train a spelling correction model based on RNN with attention~\cite{bahdanau2014neural} to correct the output of Listen, Attend and Spell (LAS) model. In contrast to this work, our model is based on Transformer architecture~\cite{vaswani2017attention} and does not require a complementary text-to-speech model for training data generation.

Transformers used for NMT are usually trained on millions of parallel sentences and tend to easily overfit if the data is scarce, which is the case we have. To solve this problem, we propose two self-complementary regularization techniques. First, we augment training data with the perturbed outputs of several ASR models trained on K-fold partitions of the training dataset. Second, we initialize both encoder and decoder with the weights of pre-trained BERT~\cite{devlin2018bert} language model, which was shown to be efficient for transfer learning in various natural language processing tasks.

We evaluate the proposed approach on LibriSpeech dataset~\cite{panayotov2015librispeech} and use Jasper DR 10x5~\cite{li2019} as our baseline ASR module. Our correction model, when applied to the greedy output of Jasper ASR, outperforms both the baseline and re-scoring with 6-gram KenLM language model and almost matches the performance of re-scoring with more powerful Transformer-XL language model.

%% file: text/2_model.tex
\section{Model}
\label{sec:model}

\subsection{Speech recognition baseline model}
\label{ssec:model}

As our baseline ASR model we use Jasper~\cite{li2019}, a deep convolutional E2E model. Jasper takes as input mel-filter bank features calculated from $20$ms windows with a $10$ms overlap and maps them to a probability distribution over characters per frame. The model is trained with Connectionist Temporal Classification (CTC) loss \cite{graves2006}.
In particular, we build on Jasper-DR-10x5, which consists of $10$ blocks of $5$ sub-blocks (1-D convolution, batch norm, ReLU, dropout) where the output of each block is added to the inputs of all following blocks similar to DenseNet~\cite{huang2017densely}. 

The baseline Jasper model is trained with the Novograd~\cite{ginsburg2019stochastic} optimizer and implemented in PyTorch within NeMo toolkit~\cite{kuchaiev2019nemo}.

\begin{figure}[ht!]
\centering
\includegraphics[width=0.8\linewidth]{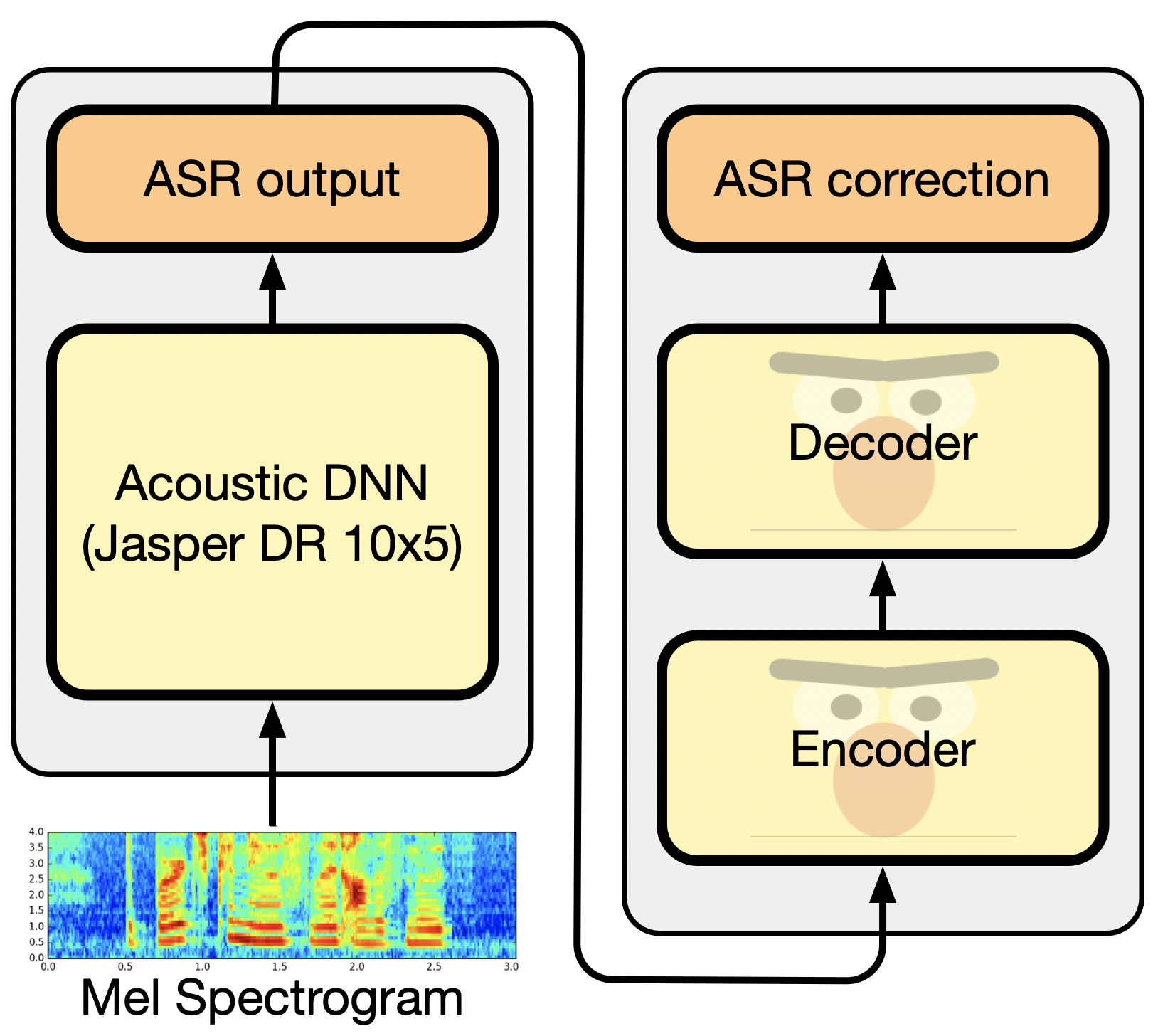}
\caption{ASR correction model based on Transformer encoder-decoder architecture.}
\label{fig:model}
\end{figure}

\subsection{Language models used for re-scoring}

A language model (LM) estimates the joint probability of a text corpus $(x_1,\dots,x_T)$ by factorizing it with a chain rule $P(x_1,\dots,x_T)=\prod_{t=1}^T P(x_t|x_1,\dots,x_{t-1})$ and sequentially modeling each conditional term in the product. To simplify modeling, it is often assumed that the context size (a number of preceding words) necessary to predict each word $x_t$ in the corpus is limited to $N$: $P(x_t|x_1,\dots,x_{t-1}) \approx P(x_t|x_{t-N},\dots,x_{t-1})$. This approximation is commonly referred to as N-gram LM.

Following the original Jasper paper~\cite{li2019}, we are considering two different LMs: 6-gram KenLM~\cite{heafield2011kenlm} and Transformer-XL~\cite{dai2019transformer} with the full sentence as the context. For generation, a beam search with the width $2048$ is used where each hypothesis is evaluated with shallow fusion of acoustic and language models:

\begin{equation}\label{eq:asr_score}
    \mathbf{y^*} = \arg\max_{\mathbf{y}} \left[ \log P_{ASR}(\mathbf{y}|\mathbf{x}) + \lambda \log P_{LM}(\mathbf{y}) \right]
\end{equation}

It is worth noting that Transformer-XL does not replace KenLM but complements it. Beam search in hypothesis formation is governed by the joint score of ASR and KenLM, and the resulting beams are additionally re-scored with Transformer-XL in a single forward pass. Using Transformer-XL instead of KenLM all the way along is too expensive due to much slower inference of the former.

\subsection{ASR correction model}

The proposed model (\Cref{fig:model}) has Transformer encoder-decoder architecture~\cite{vaswani2017attention} commonly used for neural machine translation. We denote the number of layers in encoder and decoder as $L$, the hidden size as $H$, and the number of self-attention heads as $A$. Similar to prior work~\cite{vaswani2017attention,devlin2018bert}, the fully-connected inner-layer dimensionality is set to $4H$. Dropout with probability $P_{drop}$ is applied after the embedding layer, after each sub-layer before the residual connection, and after multi-head dot-product attention.

We consider two options for initializing the model weights: random initialization and using the weights of pre-trained BERT~\cite{devlin2018bert}. Since BERT has the same architecture as Transformer encoder, its parameters can be straightforwardly used for encoder initialization. In order to initialize the decoder, which has an additional encoder-decoder attention block in each layer, we duplicate and use the parameters of the corresponding self-attention block.


%% file: text/3_experiments.tex
\section{Experiments}
\label{sec:experiments}

\subsection{Dataset}
\label{ssec:experiments_dataset}

We conduct our experiments on LibriSpeech~\cite{panayotov2015librispeech} benchmark. Librispeech training dataset consists of three parts --- train-clean-100, train-clean-360, and train-clean-500, which together provide $960$ hours of transcribed speech or around $281$K training sentences. For evaluation, LibriSpeech provides $2$ development datasets (dev-clean and dev-other) and $2$ test datasets (test-clean and test other). We found that even baseline models made only a few mistakes on dev-clean and selected the checkpoint with the lowest WER on dev-other for evaluation.

To generate training data for our Transformer ASR correction model, we split all training data into $10$ folds and trained $10$ different Jasper models in a cross-validation manner: each model was trained on $9$ folds and used to generate greedy ASR predictions for the remaining $10$th fold. Then, we concatenated all resulting $10$ folds and used Jasper greedy predictions as the source side of our parallel corpora with ground truth transcripts as the target side.

However, we did not manage to considerably improve upon Jasper greedy when training the Transformer on resulting $281$K training sentences because of extreme overfitting. To augment our training dataset, we used two techniques:
\begin{itemize}
    \item We took pre-trained Jasper model and enabled dropout during inference on training data. This procedure was repeated multiple times with different random seeds.
    \item We perturbed training data with Cutout~\cite{devries2017improved} by randomly cutting small rectangles out of the spectrogram, which essentially drops complete phonemes or words and mel frequency channels.
\end{itemize}
After the augmentation, deduplication, and removal of sentence pairs with WER greater than $0.5$, we ended up with approximately $2.5$M of training examples.

The ablation study of the proposed data augmentation techniques is presented in~\Cref{tab:baseline_data}. In our experiments, adding sentences generated with enabled dropout and cutout was much more efficient; thus, we stick to it as our training dataset in all subsequent experiments. We also experimented with using top-k beams obtained with beam search but found that the resulting sentences lacked in diversity, often differing in a few characters only. 


\input{tables/baseline_data.tex}

Recently, several promising data augmentation techniques were introduced for both NLP and ASR, such as adding noise to the beams~\cite{edunov2018understanding,zhang2019bridging} and SpecAugment~\cite{park2019specaugment} which are also applicable in our case. However, it goes beyond the scope of this paper, and we leave it for future work.

\subsection{Training}
\label{ssec:experiments_training}

All models used in our experiments are Transformers with parameters ($H=768, A=12, P_\text{drop}=0.25, L=12$). We train them with NovoGrad~\cite{ginsburg2019stochastic} optimizer ($\text{lr}=0.001, \beta_1=0.95, \beta_2=0.25$) for a maximum of $300$K steps with polynomial learning rate decay on a batches of $32$K source and target tokens. For our vocabulary we adopted $30$K WordPieces~\cite{wu2016google} used in BERT~\cite{devlin2018bert} so we can straightforwardly transfer its pre-trained weights. According to~\cite{vaswani2017attention}, we also used label smoothing of $0.1$ for regularization. Each model was trained on a single DGX-1 machine with $8$ NVIDIA V$100$ GPUs. Models were implemented in PyTorch within NeMo toolkit\footnote{\url{https://github.com/nvidia/nemo}}.

\subsection{Initialization}
\label{ssec:experiments_initialization}

Next, we experiment with various architectures and initialization schemes. Specifically, we either initialize all weights randomly (\texttt{rand}) from $\mathcal{N}(0,0.02)$ or transfer weights from the pre-trained \texttt{bert-base-uncased} model (\texttt{BERT}). \Cref{tab:models} depicts the performance of different configurations.

\input{tables/models.tex}

Models with randomly initialized encoder improve upon the results of Jasper greedy on ``other'' portions of evaluation datasets; however, their correction harms the performance on the ``clean'' portion. Adding BERT-initialized decoder achieves slightly better results, but it still lags behind the baseline Jasper with LM re-scoring.

\begin{figure}[ht!]
\centering
\includegraphics[width=\linewidth]{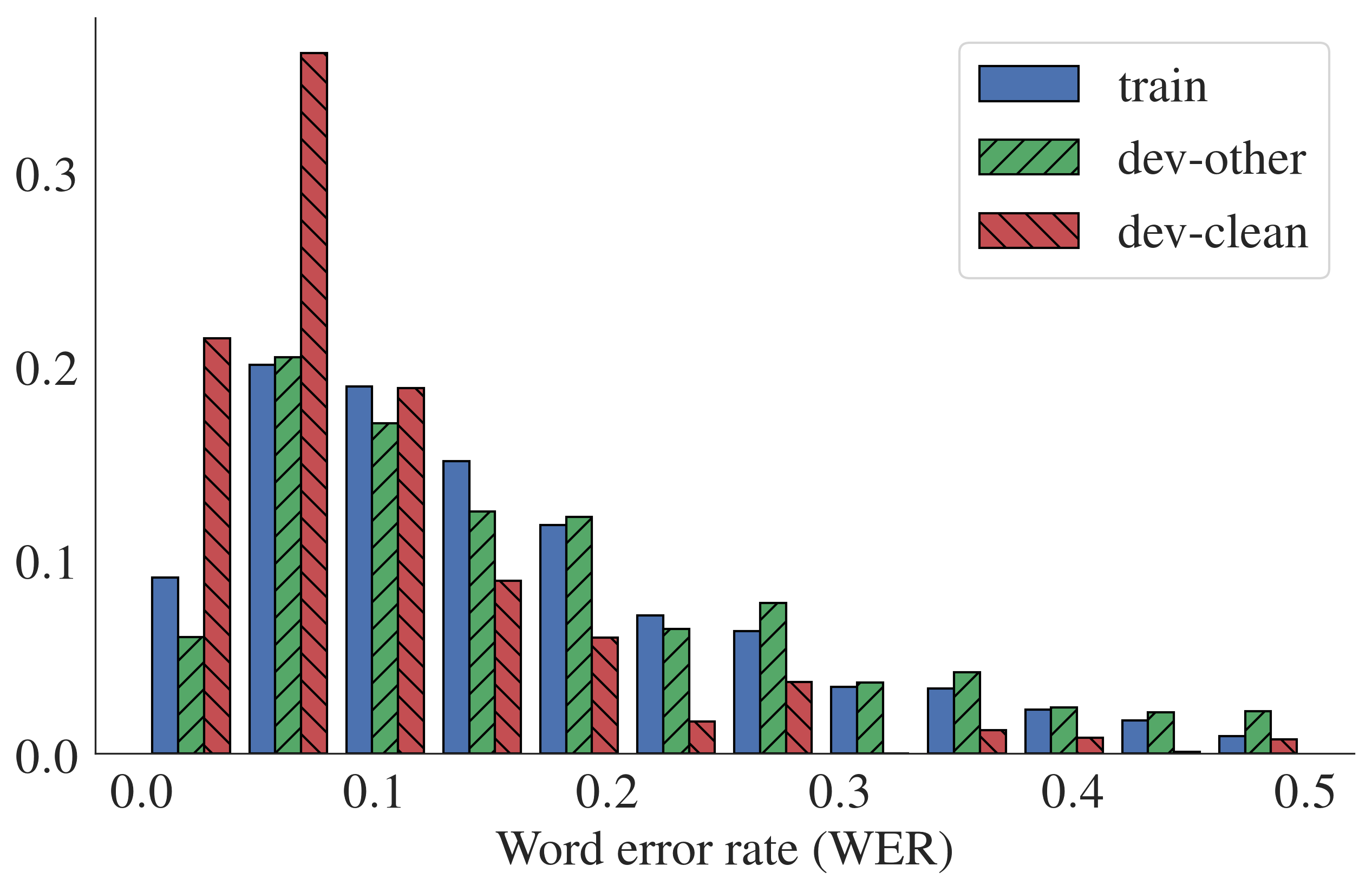}
\caption{WER distribution of training and evaluation datasets.}
\label{fig:hist}
\end{figure}

\input{tables/asr_analysis.tex}
\input{tables/correction_example.tex}

Models with BERT-initialized encoder are strictly better than both Jasper greedy and models with encoder initialized randomly. They outperform Jasper with 6-gram KenLM on ``other'' portions of evaluation datasets and approach the accuracy of the Jasper with Transformer-XL re-scoring. The best performance is achieved by the model with both encoder and decoder initialized with pre-trained BERT.

Interestingly, our ASR correction model considerably pushes forward the performance on much noisier ``other'' evaluation datasets and only moderately improves the results on ``clean''. This can be explained by the slight domain mismatch in our training data and ``clean'' evaluation datasets. Our data was collected with the models which achieve around $14\%$ WER on average (or even higher, if dropout is on during generation) and does not contain many ``clean'' examples, which usually have much lower WER. \Cref{fig:hist} shows that the distribution of WER in training data is indeed much closer to dev-other than to dev-clean.

\subsection{Analysis of corrected examples}
\label{ssec:experiments_analysis}

To conduct a qualitative analysis of ASR corrections produced by our best model with BERT-initialized encoder and decoder, we examined the examples on which it successfully corrects the output of Jasper greedy.

\Cref{tab:analysis}, which depicts the excerpts with the largest difference in WER between greedy and corrected ASR outputs, reveals an interesting pattern. Both greedy decoding and re-scoring with external LMs make a mistake at the very beginning of the speech. If it is poorly decoded by acoustic model and has little or no context for LM, the scores used for evaluating beams in \Cref{eq:asr_score} are simply unreliable. The mistakenly generated context might also negatively affect the succeeding left-to-right LM scores leading to even more errors (\Cref{tab:long_example}). Our model, on the other hand, successfully corrects ASR output by leveraging the bidirectional context provided by corrupted yet complete decoded excerpt.

%% file: tables/baseline_data.tex
\begin{table}[ht]
\centering
\begin{tabular}{lccccc} 
 \toprule
  \multirow{2}{*}{\textbf{Dataset}} & \multirow{2}{*}{\textbf{Size}} & \multicolumn{2}{c}{\textbf{Dev}} & \multicolumn{2}{c}{\textbf{Test}} \\
  & &  {\textbf{clean}} & {\textbf{other}} & {\textbf{clean}} & {\textbf{other}} \\
\midrule
 Jasper greedy & -- & $3.64$ & $11.89$ & $3.86$ & $11.95$ \\
\midrule
 10-fold & $281$K & $3.55$ & $9.70$ & $3.81$ & $9.96$ \\
 + cutout & $1.1$M & $3.35$ & $9.56$ & $3.75$ & $9.79$ \\
 + dropout & $1.7$M & $3.31$ & $9.20$ & $3.81$ & $9.52$ \\
 + both & $2.5$M & $3.26$ & $9.01$ & $3.54$ & $9.34$ \\
\bottomrule
\end{tabular}
\caption{\label{tab:baseline_data}Ablative study of data augmentation techniques.}
\end{table}

%% file: tables/models.tex
\begin{table}[ht]
\centering
\begin{tabular}{llcccc} 
 \toprule
  \multicolumn{2}{c}{\textbf{Model}} & \multicolumn{2}{c}{\textbf{Dev}} & \multicolumn{2}{c}{\textbf{Test}} \\
  {\textbf{encoder}} & {\textbf{decoder}} & {\textbf{clean}} & {\textbf{other}} & {\textbf{clean}} & {\textbf{other}} \\
\midrule
\multicolumn{2}{c}{Jasper greedy} & $3.64$ & $11.89$ & $3.86$ & $11.95$ \\
\multicolumn{2}{c}{Jasper + 6-gram} & $2.89$ & $9.53$ & $3.34$ & $9.62$ \\
\multicolumn{2}{c}{Jasper + TXL} & $2.68$ & $8.62$ & $2.95$ & $8.79$ \\
\midrule
 rand & rand & $3.92$ & $10.30$ & $4.22$ & $10.63$ \\
 rand & BERT & $3.89$ & $9.92$ & $4.19$ & $10.29$ \\
 BERT & rand & $3.26$ & $9.01$ & $3.54$ & $9.34$ \\
 BERT & BERT & $3.18$ & $8.98$ & $3.50$ & $9.27$ \\ 
\bottomrule
\end{tabular}
\caption{\label{tab:models}Performance of our model with different initialization schemes in comparison to the baselines. Jasper results are taken from the original paper~\cite{li2019}.}
\end{table}

%% file: tables/asr_analysis.tex
\begin{table*}[ht]
\centering
\begin{tabular}{llll} 
 \toprule
 {\textbf{Model}} & {\textbf{Example 1}} & {\textbf{Example 2}} & {\textbf{Example 3}}\\
\midrule
 Ground truth & pierre looked at him in surprise & i've gained fifteen pounds and & and how about little avice caro \\
\midrule
 Greedy & \underline{pure locat e ham} in a surprise & \underline{afgain} fifteen pounds and & and \underline{hawbout} little \underline{ov his carrow} \\
 + 6-gram & \underline{pure locate} him in surprise & \underline{again} fifteen pounds and & and \underline{\quad} about little \underline{of his care} \\ 
 + TXL & \underline{pure locate} him in surprise & \underline{again} fifteen pounds and & and \underline{\quad} \underline{but} little \underline{of his care} \\ 
 \textbf{Ours} & \textbf{pierre looked at him} in surprise &\textbf{i've gained} fifteen pounds and & and \textbf{how about} little \underline{of his care}\\
\bottomrule
\end{tabular}
\caption{\label{tab:analysis}Outputs produced by different models. Both greedy decoding and re-scoring with external LMs fail to recognize the beginning of the speech which is poorly decoded by acoustic model and has little or no context for LM. Our model succeeds by leveraging the context of corrupted yet complete decoded excerpt.}
\end{table*}

%% file: tables/correction_example.tex
\begin{table*}[ht]
\centering
\begin{tabular}{ll} 
 \toprule
 {\textbf{Model}} & {\textbf{Example}} \\
\midrule
 Ground truth & one day the traitor fled with a teapot and a basketful of cold victuals \\
\midrule
 Greedy & one day the \underline{trade of} fled with \underline{he teappot} and a basketful of cold \underline{victures} \\
 + 6-gram & one day the \underline{trade of} fled with  \underline{the tea pot} and a basketful of cold \underline{pictures} \\ 
 + TXL & one day the \underline{trade of} fled with  \underline{the tea pot} and a basketful of cold \underline{victores} \\ 
\textbf{Ours} & one day the \underline{trader} fled with \textbf{a teapot} and a basketful of cold \textbf{victuals} \\
\bottomrule
\end{tabular}
\caption{\label{tab:long_example} Combination of acoustic and language models fails to generate the subject of the sentence which leads to further errors. While our ASR correction model does not manage to fully reconstruct the ground truth transcript, its output is coherent English with the last word successfully corrected.}
\end{table*}

%% file: text/4_conclusion.tex
\section{Conclusion}
\label{sec:conclusion}

In this work, we investigated the use of Transformer-based encoder-decoder architecture for the correction of ASR systems output. The proposed ASR output correction technique is capable of ``translating'' the erroneous output of the acoustic model into grammatically and semantically correct text.

The proposed approach enhances the acoustic model accuracy by a margin comparable to shallow fusion and re-scoring with external language models. Analysis of corrected examples demonstrated that our model works in scenarios when the scores produced by both acoustic and external language models are not reliable.

To overcome the problem of extreme overfitting on the relatively small training dataset, we proposed several data augmentation and model initialization techniques, i.e., enabling dropout and Cutout~\cite{devries2017improved} during acoustic model inference and initializing both encoder and decoder with the parameters of pre-trained BERT~\cite{devlin2018bert}. We also performed a series of ablation studies showing that both data augmentation and model initialization have a significant impact on model performance.
